\documentclass{article}

\PassOptionsToPackage{numbers}{natbib}
\usepackage{ml_institute}

\usepackage{booktabs}
\usepackage[table]{xcolor}
\usepackage{multirow}
\usepackage{xcolor}
\usepackage{subcaption}
\usepackage{colortbl}
\usepackage{url}
\usepackage{float}
\usepackage{multirow}
\usepackage{booktabs} 
\usepackage{amssymb}
\usepackage{amsmath}
\usepackage{bm}
\usepackage{graphicx}
\usepackage{makecell}
\usepackage{arydshln}
\usepackage{booktabs}
\usepackage{array}
\usepackage{booktabs}
\usepackage{graphicx}

\definecolor{cbred}{rgb}{0.9255, 0.6157, 0.5922}
\definecolor{cbgreen}{rgb}{0.2275, 0.6863, 0.4275}

\definecolor{deemph}{gray}{0.6}

\usepackage{pifont}

\definecolor{defaultcolor}{rgb}{0.8666, 0.8666, 0.8666}

\usepackage{float}

\title{Occlusion-Aware 3D Motion Interpretation for Abnormal Behavior Detection}

\author{
Su Li$~^{1,2}$ \quad
Liang Wang$~^{2}$ \quad
Jianye Wang$~^{2}$ \quad \\ \bf
Ziheng Zhang$~^{2}$ \quad
Lei Zhang$~^{3}$\\ \\
$~^{1}$~University of Science and Technology of China, Hefei, China\\
$~^{2}$~Hefei Institutes of Physical Science, Chinese Academy of Sciences, Hefei, China\\
$~^{3}$~Laboratory of Vision Engineering (LoVE),\\
 School of computer science, 
 University of Lincoln, Lincoln, UK\\
 \texttt{lisu@mail.ustc.edu.cn}
}

\def\etal{\emph{et al. }}

\begin{document}

\maketitle

\begin{abstract}%

Estimating abnormal posture based on 3D pose is vital in human pose analysis, yet it presents challenges, especially when reconstructing 3D human poses from monocular datasets with occlusions. Accurate reconstructions enable the restoration of 3D movements, which assist in the extraction of semantic details necessary for analyzing abnormal behaviors. However, most existing methods depend on predefined key points as a basis for estimating the coordinates of occluded joints, where variations in data quality have adversely affected the performance of these models. In this paper, we present OAD2D, which discriminates against motion abnormalities based on reconstructing 3D coordinates of mesh vertices and human joints from monocular videos. The OAD2D employs optical flow to capture motion prior information in video streams, enriching the information on occluded human movements and ensuring temporal-spatial alignment of poses. Moreover, we reformulate the abnormal posture estimation by coupling it with Motion to Text (M2T) model in which, the VQVAE is employed to quantize motion features. This approach maps motion tokens to text tokens, allowing for a semantically interpretable analysis of motion, and enhancing the generalization of abnormal posture detection boosted by Language model. Our approach demonstrates the robustness of abnormal behavior detection against severe and self-occlusions, as it reconstructs human motion trajectories in global coordinates to effectively mitigate occlusion issues. Our method, validated using the Human3.6M, 3DPW, and NTU RGB+D datasets, achieves a high $F_1-$Score of 0.94 on the NTU RGB+D dataset for medical condition detection. And we will release all of our code and data. \\
\end{abstract}

\section{Introduction}

Abnormal behavior detection is critical for enhancing security, healthcare, and sports applications. In this task, accurately identifying abnormal movements depends on precise, which involves reconstructing 3D human poses motions from monocular datasets. Nevertheless, significant challenges remain, including the ambiguous nature of defining an anomaly, the scarcity of abnormal data, difficulties in reconstructing 3D poses from 2D frames, and particularly, occlusions that obscure movements. To some extent, this task relies on advancements in 3D Human Pose Estimation (3DHPE) research.

\begin{figure*}[t]
 \centerline{\includegraphics[width=1.0\textwidth]{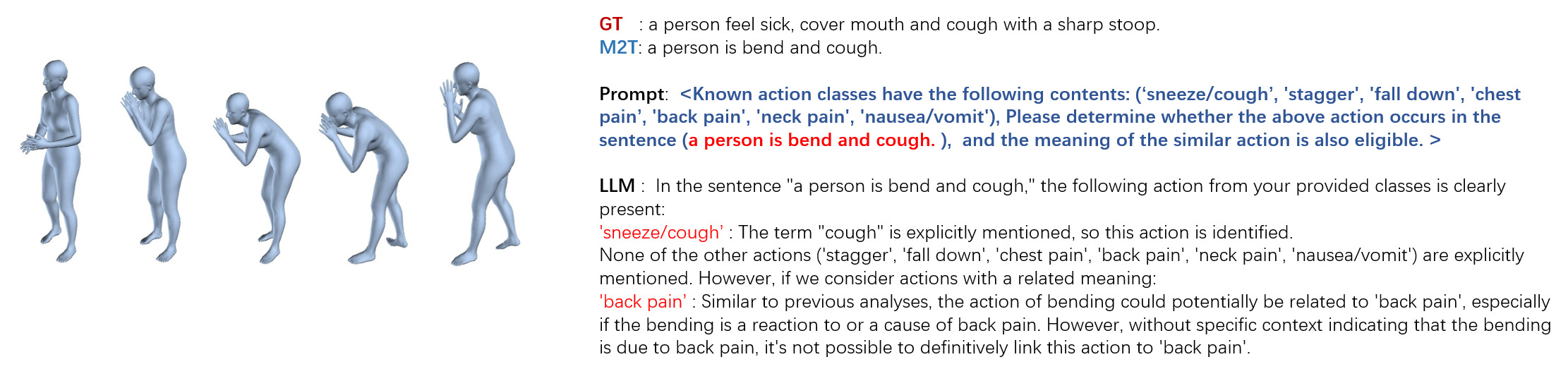}}
\caption{The M2T model captures motion semantics, and LLM can use specifical prompts identify behavior classes in medical conditions.}
\label{fig:prompt}
\end{figure*}

3DHPE using a monocular camera directly regresses from RGB images to 3D human poses, utilizing single frames~\cite{kanazawa2018end, kolotouros2019learning, pavlakos2018learning} or video sequences~\cite{kanazawa2019learning, kocabas2020vibe, luo20203d}, employs deep Convolutional Neural Networks (CNNs) to predict 3D pose joints directly~\cite{kocabas2021pare, wehrbein2021probabilistic} or predict human pose parameters like those of the Skinned Multi-Person Linear model (SMPL)~\cite{loper2023smpl}. While achieving high accuracy on various indoor datasets, challenges like severe occlusions can disrupt model performance due to incorrect parameter regression and this misalignment may result in erroneous model outputs. To counteract occlusion, several studies~\cite{xie2021physics, tripathi20233d} suggest incorporating priori physical knowledge into the regression process, enhancing robustness. Current studies are focused on spatio-temporal modeling~\cite{arnab2019exploiting, liu2021graph}, which utilize motion data to accurately depict human movements and handle occlusions effectively. These efforts focus on creating a robust human motion model that handles occlusions in current frames by learning implicit cues from contextual sequences. However, these models heavily rely on static features to ensure temporally consistent and smooth representations of 3D human body motion across video sequences~\cite{choi2021beyond}. Techniques like optical flow estimation~\cite{ding2020every, shi2023videoflow} infer motion in occluded areas from adjacent pixels, aiding accuracy but still struggling to balance overfitting in typical scenes against underfitting in complex scenarios. Furthermore, existing methods do not adequately capture semantic information of abnormal actions, highlighting a gap in effectively recognizing such behaviors.

\begin{figure*}[t]
\centering
\subfloat[2D Frames to 3D Poses.]{
    \label{fig:2_a}
    \includegraphics[width=0.985\textwidth]{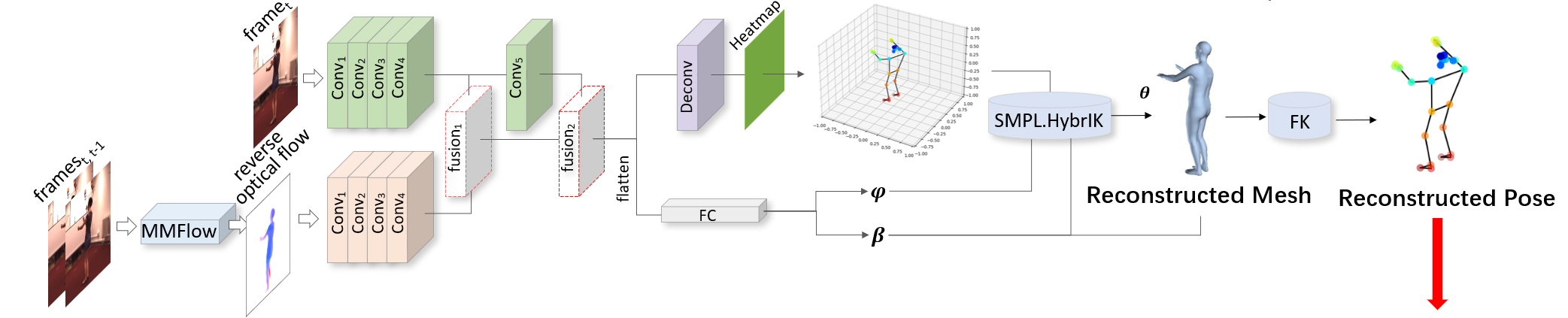}
}
\\
\subfloat[Interpretable Anomaly Detection.]{
    \label{fig:2_b}
    \includegraphics[width=1.0\textwidth]{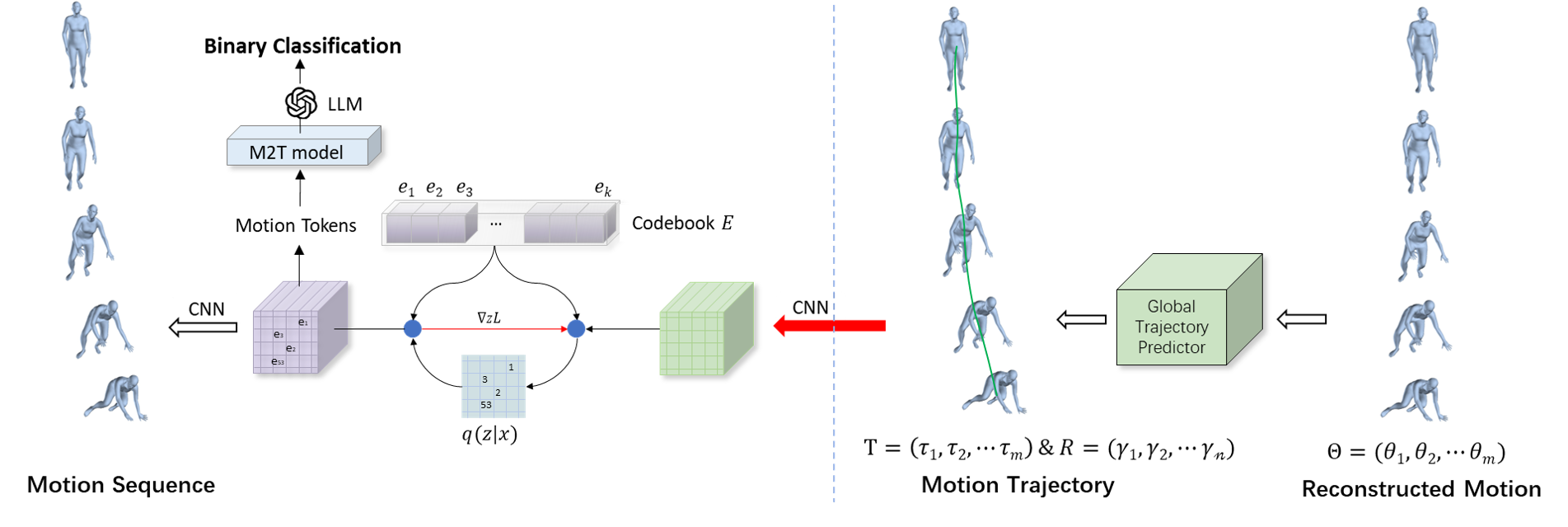}
}
\caption{\textbf{Overview of our method (OAD2D).} The entire pipeline consists of pose estimation using image and optical flow, followed by trajectory optimization and motion quantization, along with LM classification. (a) A two-stream neural network generates 3D heatmaps to facilitate regression to the 3D joints, and, when integrated with kinematics, it outputs shape parameters $\beta$, twist angles $\varphi$, and pose $\theta$ for motion reconstruction ($\Theta$). (b) The global trajectory predictor creates corresponding global trajectories $(T, R)$, which include root translation $T$ and rotation $R$. This motion data is then quantized by a VQVAE to construct the codebook $E$, motion tokens, and subsequently to generate semantic motion representations for abnormal behavior detection through sentiment analysis via LLM.}
\label{fig:2}
\end{figure*}

Building on the insights from previous research, in this article, we propose OAD2D ({\bf O}cclusion-Aware {\bf A}bnormality {\bf D}etection in {\bf 2D} Frames), a novel method for detecting abnormal actions in occlusion. The overall pipeline is shown in Fig.~\ref{fig:2}. Initially, we employ optical flow~\cite{2021mmflow, jiang2021learning} for temporal and spatial alignment, as well as for compensating occlusion, to facilitate the regression of keypoint heatmaps. Subsequently, to accurately reconstruct human behavior understanding in 3D pose from monocular video, recovering the motion trajectory under global coordinates~\cite{yuan2022glamr} is crucial. To this end, we enhance motion information using physical constraints~\cite{li2021hybrik} and a global coordinate transformation, thereby achieving a more precise motion representation under occlusion. Finally, based on the reconstructed 3D pose, we reconceptualize abnormal behavior detection by treating motion as a language, utilizing the Motion-to-Text (M2T) model~\cite{radouane2024motion2language} to translate motion tokens into text tokens for semantic interpretation. Specifically, we employ a Vector Quantized Variational Autoencoder (VQVAE)~\cite{van2017neural} to compress 3D motion data, encapsulating motion features as discrete tokens within a codebook. The  experiments demonstrate that the effectiveness of detecting pose anomalies in occlusion scenarios using our method and it also allows for semantic interpretation of abnormal behaviors. By integrating with large language models (LLMs)~\cite{achiam2023gpt} for few-shot learning, we can achieve better performance than purely visual models, providing alternative direction for interpretable, as shown in Fig. \ref{fig:prompt}.

The contributions of this paper are as follows: 
(1) We propose a new approach named OAD2D, which sets a new benchmark for accurately detecting abnormal actions in scenarios with occlusion. 
(2) We employ a novel 3D pose regressor to facilitate 2D-to-3D pose regression under occlusion along with a global trajectory predictor to enhance global trajectory prediction for the generated poses.
(3) We pioneer the use of LLM in conjunction with our occlusion-aware detection system to provide semantic interpretation of detected motions. 

\vspace{1cm}

\section{Related Work}

The challenge of estimating 3D human pose and shape from monocular video~\cite{wang2021deep, ling2023dl3dv, kocabas2020vibe} has propelled the development of several advanced methods. These approaches, grounded in model-based techniques~\cite{kanazawa2018end, kolotouros2019learning}, leverage temporal information and advanced neural network architectures~\cite{choi2021beyond, 9320338} to enhance pose estimation. Choi \etal introduced the Time Consistent Mesh Restoration System (TCMR)~\cite{choi2021beyond}, which employs bidirectional GRUs to focus on temporal features, while others like PoseNet3D~\cite{9320338} utilize knowledge distillation from 2D skeletons to estimate SMPL~\cite{loper2023smpl} parameters without requiring 3D training data. Attention mechanisms and hybrid inverse kinematics~\cite{li2021hybrik} solutions further improve the accuracy of pose reconstruction, accounting for kinematic constraints and depth ambiguities.

In parallel, advancements in optical flow estimation~\cite{2021mmflow, lu2023transflow}, such as the FlowNet~\cite{dosovitskiy2015flownet} series and PWC Net~\cite{sun2018pwc}, have deepened our understanding of pixel motion, contributing to more stable and accurate motion predictions. These innovations in optical flow techniques are enriched by the integration of semantic segmentation and motion propagation modules, enhancing the robustness and consistency of flow estimation in dynamically complex environments.

Representing motion features in videos has been another focus, which involves capturing motion dynamics such as trajectory~\cite{yuan2022glamr} and velocity using methods like dense optical flow and vector fields. The pre-training and fine-tuning stages of MotionBERT~\cite{zhu2023motionbert}, along with modular designs like FrankMocap~\cite{rong2020frankmocap}, help in creating more sophisticated and detailed 3D motion feature representations. Moreover, recent studies advocate for learning discrete motion representations using VQVAE~\cite{van2017neural}, enabling the generation and mapping of complex human motion patterns to textual descriptions. 

For the abnormal behavior recognition methods~\cite{sultani2018real, ji2024algorithm, hao2021end}, they principally rely on classifiers of visual cue and features, powered by deep learning networks. However, these works, anchored in 2D features for human behavior anomaly detection, lack interpretability and heavily depend on precise anomaly label information. Furthermore, semantic ambiguity significantly hampers abnormal behavior discrimination, particularly in medical conditions.

\section{Proposed Method}

The pipeline of our proposed method is illustrated in Fig.~\ref{fig:2}. In Fig.~\ref{fig:2_a}, the two-stream network is trained on both video frames and their optical flow to generate 3D poses even in the presence of occlusion (Sec.~\ref{sec:generator}). Subsequently, based on the generated poses shown in Fig.~\ref{fig:2_b}, global trajectory prediction (Sec.~\ref{sec:predictor}) is completed via CVAE and fed into VQVAE and reference to generate motion tokens. These tokens are then utilized to derive semantic information using the M2T model. Finally, abnormal behaviors are discriminated by LLMs (Sec.~\ref{sec:vqvae}).

\subsection{3D Pose Regressor with Occlusion}
\label{sec:generator}
Some prior researches~\cite{rong2021frankmocap, kocabas2021pare} often experience missing 2D joints resulting due to occlusion, subsequently hindering accurate human mesh inference. Building upon previous works in kinesiology~\cite{xu2020deep,xie2021physics,li2021hybrik}, we have developed a 3D human pose generator that fuses optical flow with kinematics. This integration not only alleviates the impact of occlusions on human reconstruction but also enhances regression of human shape parameters. Inspired by HybrIK~\cite{li2021hybrik}, in our method, the pose is estimated through direct joint regression, complemented by the inverse kinematics with SMPL~\cite{loper2023smpl}.

\textbf{3D Keypoint Regression.} We first utilize MMFlow~\cite{2021mmflow} to infer the corresponding optical flow images for video sequences. After generating a 3D heatmap through deconvolution layers followed by a $1\times1$ convolution kernel, we apply soft-argmax to extract the 3D poses from the heatmap. The keypoint locations are then predicted under the supervision of ${L}1$ loss:

\begin{equation}
\label{eq:joints_loss}
      {L}_{keypoints}=\frac{1}{K} \sum_{k=1}^{K}\left\|p_{k}-\hat{p}_{k}\right\|_{1},
\end{equation}
where $K$ is the total number of keypoints and $\hat{p}_{k}$ is the ground-truth 3D keypoint.
\vspace{0.1cm}

\textbf{SMPL Parameters Regression.} Upon generating the twist angles $\varphi$ and shape parameters $\beta$ through the fully-connected layer, hybrid analytical-neural inverse kinematics is then utilized to determine the relative rotation(i.e. the pose parameters $\theta$) of the 3D poses.The twist angles $\varphi$ are supervised using ${L}2$ loss: 

\begin{equation}
\label{eq:twist_loss}
      {L}_{twist}=\frac{1}{K} \sum_{k=1}^{K}\left\|\left(\cos{\varphi_{k}}, \cos{\varphi_{k}}\right)-\left(\cos \hat{\varphi}_{k}, \sin \hat{\varphi}_{k}\right)\right\|_{2},
\end{equation}
where $\hat{\varphi}_{k}$ is the ground-truth twist angle.

We use these parameters put into the SMPL model to reconstruct the human mesh. During the training phase, the shape parameters $\beta$ and pose parameters $\theta$ of SMPL are supervised as follows:

\begin{equation}
      {L}_{shape}=\|\beta-\hat{\beta}\|_{2}\text{,} \quad {L}_{pose}=\|\theta-\hat{\theta}\|_{2},
\label{eq:smpl_loss}
\end{equation}

\subsection{Global Trajectory Predictor}
\label{sec:predictor}

When the 3D pose regressor is used to estimate the pose $\theta$ with occlusion, it struggles to accurately depict the direction, trajectory, and global coordinates of human motion due to the absence of root information. In this case, we utilize the Global Trajectory Predictor $\mathcal{T}$ from GLAMR~\cite{yuan2022glamr} to take the pose $\Theta = (\theta_1,\ldots, \theta_t)$ as input and forecast the corresponding root translations $T = (\tau_1, \ldots, \tau_t)$ and rotations $R = (\gamma_1, \ldots, \gamma_t)$. To reduce the trajectory ambiguity caused by occlusion, the Global Trajectory Predictor operates within the CVAE framework:

\begin{equation}
\label{eq:s_traj}
{\Psi}=\mathcal{T}(\Theta, v),
\end{equation}
\begin{equation}
\label{eq:g_traj}
(T, R)={EgoToGlobal}(\Psi),
\end{equation}
where $\mathcal{T}$ is the decoder of CVAE and $v$ is the latent code obtained by the pose passing through the encoder of CVAE, the output $\Psi = (\psi_1, \ldots, \psi_t)$ is the self-centered trajectory and converted to a global trajectory $(T, R)$ via the conversion function.

\subsection{Interpretable Motion Expressor}
\label{sec:vqvae}

As shown on the left in Fig. \ref{fig:2_b}, our approach leverages LLMs to detect abnormal behaviors, the powerful textual performance of LLMs derives from the Transformers~\cite{vaswani2017attention} framework. The detection of abnormal behavior can be framed as a binary classification problem (abnormal or normal), then by achieving an accurate semantic representation of human motion and using the right prompts, we can avoid relying on traditional visual classification methods~\cite{simonyan2014very, goodfellow2014generative}. This not only enriches the motion representations but also enhances their interpretability.

Guided by the aforementioned motivation, our methodology facilitates the transformation from 2D frames to 3D poses under occlusion as detailed in Sec \ref{sec:generator} and Sec. \ref{sec:predictor}. We leverage trajectory information, including the root node, and enhance it by incorporating detailed motion data—such as velocity and acceleration of the joints—referenced from the HumanML3D~\cite{Guo_2022_CVPR} data format. This approach provides a more comprehensive representation of motion information. We utilize the VQVAE~\cite{van2017neural} to convert motion into tokens by building a codebook through motion quantization and modeling motion-to-text translation.

\textbf{Motion Quantization.}To reduce data dimensionality and facilitate the replacement of complex motion data with representative features, we employ a VQVAE to learn latent codes for 3D human motion. A series of motion sequences $m \in \mathbb{R}^{T \times D_{p}}$ as input, where $T$ is the number of poses contained in all frames and $D_{p}$ is pose dimension, is processed along the temporal dimension using $1\times1$D convolution to generate latent vectors $\hat e \in \mathbb{R}^{t \times d}$, where $d$ represents the number of convolution kernels. This transformation is expressed as $\hat e=Encoder(m)$. The vectors $\hat e$ are then subjected to discrete quantization $Q$, converting them into a set of codebook entries $e_k$. The learnable codebook $E=\{{e}\}_{k=1}^{K}$ consists of $K$ potential embedded $d$-dimensional vectors. The quantization process $Q$ involves mapping each vector $\hat{e}_k$ into the nearest vector in the codebook $E$, defined as follows:

\begin{equation}
\label{eq:encoder}
    e_k=Q(\hat e):=\left(\operatorname{argmin}_{{e}_{i} \in E}\left\|\hat{e}_{i}-{e}_{k}\right\|\right) \in \mathbb{R}^{t \times d}.
\end{equation}

And then, projecting $e_k$ as a sequence of poses $\hat m$ back into the space of 3D motion through a decoder $D$, the process is expressed as:

\begin{equation}
\label{eq:decoder}
   \hat m=D(e_k)=D(Q(E(m))).
\end{equation}

The loss fucntion of VQVAE consists of a reconstruction loss, a codebook loss, and an encoder output loss:

\begin{equation}
    L_{vqvae}=\|\hat m-m\|_{1}+\left\|\operatorname{sg}[E(m)]-e_{k}\right\|_{2}^{2}+\beta\left\|E(m)-\operatorname{sg}\left[e_{k}\right]\right\|_{2}^{2},
\label{eq:loss_vq}
\end{equation}where $\beta$ is the empirical hyperparameter and $sg$ is the stop gradient.

\textbf{Motion-to-Text Model Learning.} Following the discrete quantization of motion, we can convert motion into tokens, which can then be directly used with the Motion-to-Text (M2T) model. This allows us to learn the mapping relationship between human motions and textual descriptions. The target of this mapping is the text tokens $c \in\{1, \ldots,|V|\}^{N}$ following encoding methods like Glove, where $V$ represents the vocabulary and $N$ is the number of words. In the M2T model, motion tokens are fed into the transformer encoder, and subsequently, the decoder predicts the probability distribution of potential discrete text tokens at each step $p_{\theta}(c \mid s)=\prod_{i} p_{\theta}\left(c_{i} \mid c_{<i}, s\right)$. This process defines the model's loss function as follows:

\vspace{-0.4cm}
\begin{equation}
\label{eq:loss}
    {L}_{m2t}=-\sum_{i=0}^{N-1} \log p_{\theta}\left(c_{i} \mid c_{<i}, s\right).
\end{equation}

\textbf{Abnormal Behavior Detection with LLMs.} As previously mentioned, we utilize motion tokens to map text tokens. After inference using the decoder of the transformer, we obtain the semantic representation of the motion. Subsequently, we enable LLMs to directly perform the task of detecting abnormal behavior by using a specific prompt through in context learning, as illustrated in Fig. \ref{fig:prompt}.

\section{Experiments}
\subsection{Experimental Setup}
\textbf{Datasets.} For the 3D pose regressor, we utilize the \textbf{Human3.6M}~\cite{h36m_pami} and \textbf{3DPW}~\cite{vonMarcard2018} datasets for both training and evaluation. Given that our approach is based on an optical flow driven two-stream network, we also incorporated the corresponding optical flow datas from Human3.6M and 3DPW datasets.
Regarding the detection of abnormal behaviors, we select the \textbf{NTU RGB+D 120}~\cite{liu2020ntu, shahroudy2016ntu} dataset, which features explicit movement classification. We focus on abnormal behaviors included movements associated with medical conditions (NTU-MC) as they relate to everyday life activities, so we also add \textbf{HumanML3D}~\cite{Guo_2022_CVPR} dataset as a supplement to the daily normal behavior data. 

\textbf{Implementation Details.} All experiments are conducted using PyTorch. On our two-stream model phase, building upon prior research, the pre-trained weight of image pathway are used to continue learning from 2D source data. The optical flow data inferred from Human3.6M via MMflow is utilized as the input, with all images standardized to a size of 256 × 192. This model undergo feature fusion at both mid and end stages, culminating in outputs through a fully connected layer and deconvolution layers, which is divided into  generating 3D heatmaps for 3D pose regression and regressing SMPL parameters, comprises a final layer with 56 neurons (10 for $\beta$ and 46 for $\theta$). The learning rate is set at $1 \times 10^{-3}$, with reductions by a factor of ten at both the 10th and 15th epochs. The model is optimized using Adam and trained on NVIDIA RTX 3090 GPUs over 20 epochs, with a mini-batch size of 56, across four GPUs. 
Based on VQVAE, a motion discretizer is trained to generate a codebook $E$ using a single GPU, with batch size adjustable according to memory size. The training lasted 300 epochs, initiating GAN at the 100th epoch with a learning rate of $1 \times 10^{-3}$. Due to the small frame count of NTU-MC and HumanML3D, the motion window size is set to 32 to capture motion features. The codebook size is 1024. During the motion to text phase, a transformer is trained to map motion tokens to text tokens, utilizing a single GPU with a batch size of 1200 for 400 epochs, at a learning rate of $1 \times 10^{-4}$.

\textbf{Evaluation Metric.} In the 3D pose preprocessing approach, we deploy three established metrics to evaluate pose estimation: \textbf{MPJPE} calculates average discrepancy between predicted and real joint positions in 3D human pose estimations. \textbf{PA-MPJPE} is an improved MPJPE applying Procrustes analysis to minimize translation, rotation, and scale differences before error computing, enhancing joint localization accuracy. \textbf{MPVPE} measures the average difference in 3D position between predicted and true model vertices, assessing 3D model accuracy.
In binary classification, \textbf{Accuracy} measures overall correctness, while $\mathbf{F}_1$\textbf{-Score} is the harmonic mean of precision and recall, crucial for imbalanced datasets. \textbf{Recall} gauges the model's detection of positives, and precision the accuracy of these detections. \textbf{Macro and Weighted Averages} offer insights into performance across classes, with the latter accounting for class frequency. These metrics together provide a nuanced picture of predictive performance.

\begin{table}[htbp]
    \centering
    \footnotesize
    \begin{tabular}{@{}lcccccc@{}}
    \toprule
    \multirow{2}{*}{Methods} & \multicolumn{3}{c}{3DPW-TEST} & \multicolumn{2}{c}{Human3.6M-TEST} \\ 
    \cmidrule(r){2-4} \cmidrule(r){5-6} & PA-MPJPE ($\downarrow$) & MPJPE ($\downarrow$) & MPVPE($\downarrow$)  & PA-MPJPE ($\downarrow$) & MPJPE ($\downarrow$) \\
    \midrule
    HMR~\cite{kanazawa2018end} & 76.7 & 130.0 & - & 56.8 & 88.0 \\
    SPIN~\cite{kolotouros2019learning} & 59.2 & 96.9 & 116.4 & 41.1 & 62.5\\
    Pose2Mesh~\cite{choi2020pose2mesh} & 58.3 & 88.9 & 106.3 & 46.3 & 88.9 \\
    TCMR~\cite{choi2021beyond} & 52.7 & 86.5 & 103.2 & 52.0 & 86.5\\
    VIBE~\cite{kocabas2020vibe} & 51.9 & 82.9 & 99.1 & 41.4 & 65.6\\
    [0.5ex]\hdashline\noalign{\vskip 0.5ex}
    $\operatorname{Ours}$ & \textbf{49.7}  &  \textbf{80.3}  &  \textbf{97.6} & \textbf{37.3} & \textbf{58.8} \\
    \bottomrule
\end{tabular}
\vspace{0.2cm}
\caption{\textbf{3D Human Pose Estimation:} Evaluation on 3DPW and Human3.6M. }
\label{tab:3dhpe result}
\end{table}

\subsection{Main Results}
\textbf{Quantitative Results.}
In order to verify the validity of our model in different settings, we tested our model on different datasets, i.e., 3DPW~\cite{vonMarcard2018}, Human3.6M~\cite{h36m_pami}, NTU-MC~\cite{liu2020ntu}, and HumanML3D~\cite{Guo_2022_CVPR}.

We initiate our evaluation by comparing the performance of our 3D pose regressor against various baseline models as detailed in Tab. \ref{tab:3dhpe result}. Reference to prior works~\cite{kocabas2020vibe, yuan2022glamr, li2021hybrik}, we fine-tuning our model using data from Human3.6M-TRAIN and 3DPW-TRAIN. The results on the validation sets of both datasets demonstrate our model's superior performance, evidenced by the following results: on Human3.6M-TEST, we achieved a PA-MPJPE of 37.3mm and an MPJPE of 58.8mm; on 3DPW-TEST, we recorded a PA-MPJPE of 49.7mm, an MPJPE of 80.3mm, and an MPVPE of 97.6mm.

In order to evaluate the performance of the abnormal behavior detection method through interpretable 3D motion, we evaluate the precision of detecting 12 distinct disease-related behaviors in the NTU-MC dataset, as depicted in Tab. \ref{tab:medical_conditions}. Since we are more interested in disease behaviors that express body lanuage, we derive a subset named MC-Half (See bolded classes in Tab. \ref{tab:medical_conditions}) to validate the results of our classification. The detailed evalation results in Tab. \ref{tab:ntu}.

To further validate the generalization of our method, we crafte a hybrid dataset by merging HumanML3D with NTU-MC, conscientiously maintaining a low-probability distribution for abnormal behaviors, as typically observed in everyday life. As illustrated in Tab. \ref{tab:llm}, using this approach, we achieved an accuracy of 85\% in the identification of abnormal behaviors. Moreover, the $F_1$-score for the recognition of normal behaviors reached a high mark of 0.90. 

\begin{table}[htbp]
  \centering
  \footnotesize
  \begin{tabular}{@{}lcccc@{}}
    \toprule
    Datasets & Action Classes & Accuracy & $F_{1}$-Score \\
    \midrule
    NTU-MC & 12 & 87.3\% & 0.93  \\
    MC-Half & 6 & 89.3\% & 0.94 \\
    \bottomrule
  \end{tabular}
  \vspace{0.2cm}
  \caption{\textbf{Abnormal Behavior Detection} Performance on NTU-MC and half classes of NTU-MC(MC-Half).}
  \label{tab:ntu}
  \vspace{-0.5cm}
\end{table}

\begin{table}[htbp]
\centering
\begin{minipage}[t]{0.4\linewidth}
\centering
\footnotesize
\begin{tabular}{|l|l|}
\hline
A41: sneeze/cough & \textbf{A42: staggering} \\
\hline
\textbf{A43: falling down} & A44: headache \\
\hline
\textbf{A45: chest pain} & \textbf{A46: back pain} \\
\hline
\textbf{A47: neck pain} & \textbf{A48: vomiting} \\
\hline
A49: fan self & A103: yawn \\
\hline
A104: stretch oneself & A105: blow nose \\
\hline
\end{tabular}
\vspace{0.2cm}
\caption{Medical Conditions.}
\label{tab:medical_conditions}
\end{minipage}%
\hfill
\begin{minipage}[t]{0.55\linewidth}
\centering
\footnotesize
\begin{tabular}{@{}lcccc@{}}
\toprule
 & Precision & Recall & $F_{1}$-Score & Support \\
\midrule
Normal & 0.88 & 0.92 & 0.90 & 17264  \\
Abnormal & 0.78 & 0.89 & 0.94 & 6638 \\
Accuracy & - & - & 0.85 & 23902 \\
Macro Avg & 0.82 & 0.79 & 0.81 & 23902 \\
Weighted Avg & 0.85 & 0.85 & 0.85 & 23902 \\
\bottomrule
\end{tabular}
\vspace{0.2cm}
\caption{\textbf{LLM Classification} Performance on HumanML3D and MC-Half.}
\label{tab:llm}
\end{minipage}
\end{table}

\begin{figure*}[t]
\centering
\subfloat[Frames]{
    \label{fig:a}
    \includegraphics[width=0.16\textwidth]{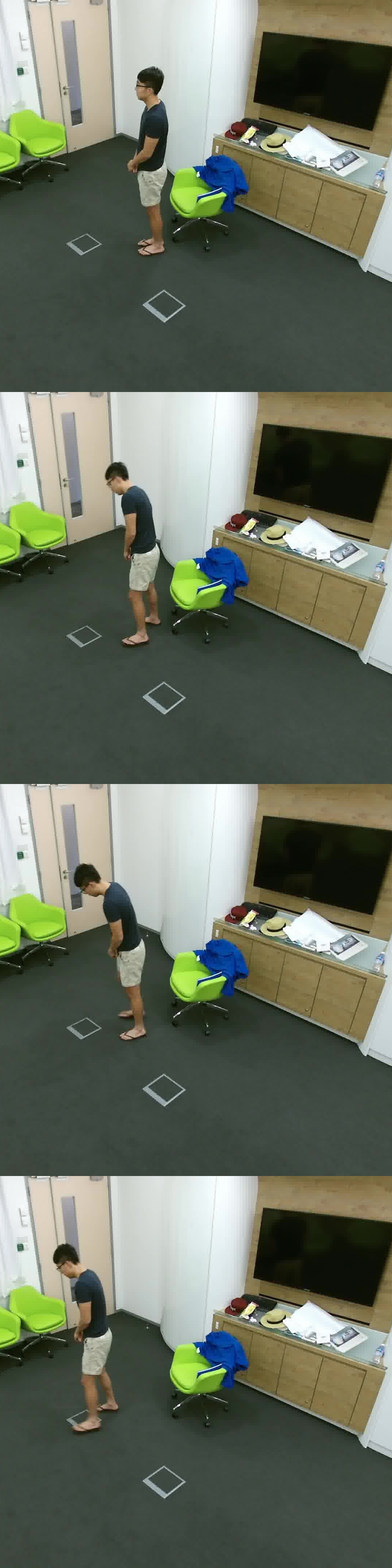}
}
\subfloat[3D Pose]{
    \label{fig:b}
    \includegraphics[width=0.16\textwidth]{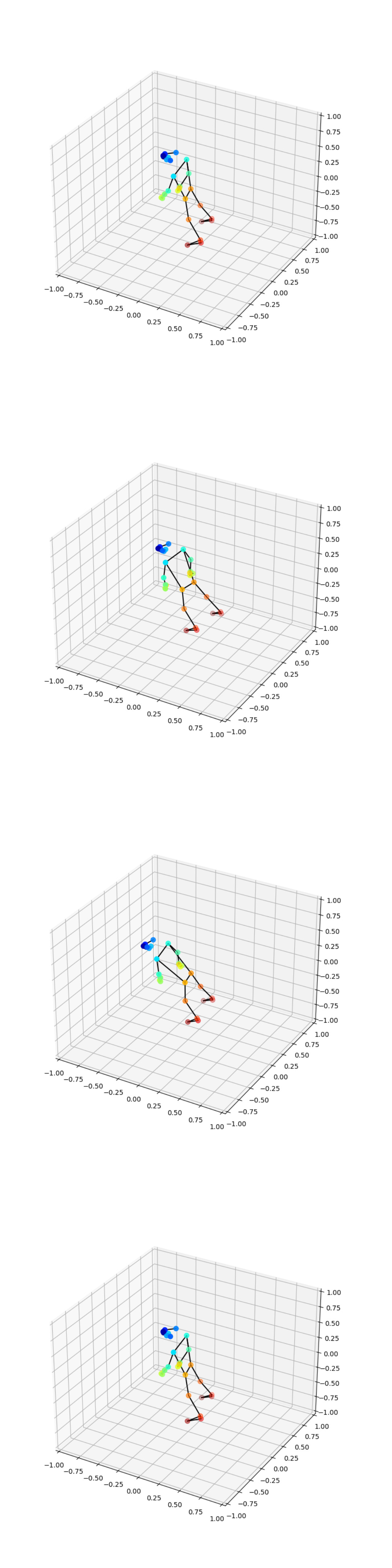}
}
\subfloat[Optical Flow]{
    \label{fig:c}
    \includegraphics[width=0.16\textwidth]{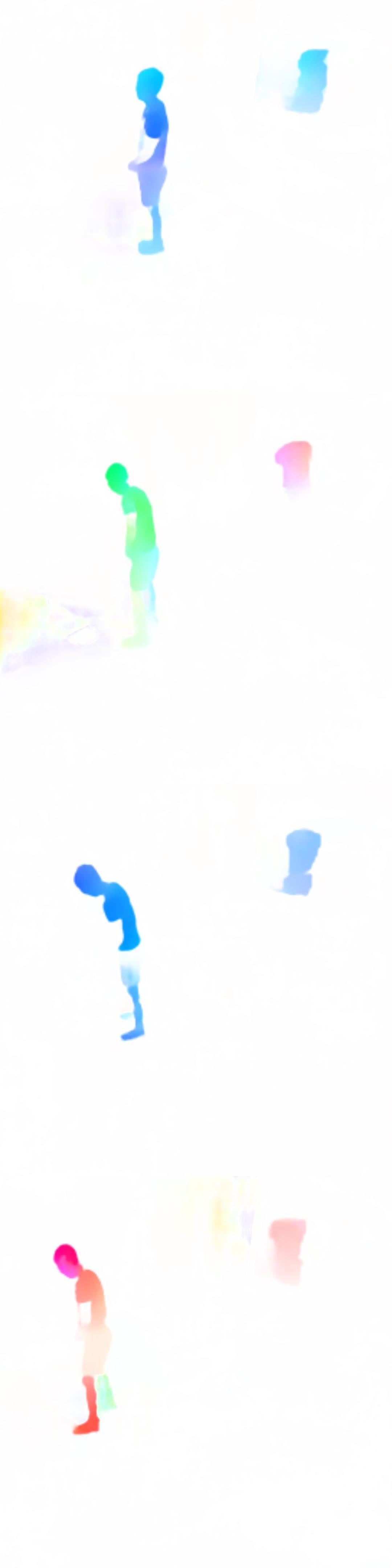}
}
\subfloat[HybrIK]{
    \label{fig:d}
    \includegraphics[width=0.16\textwidth]{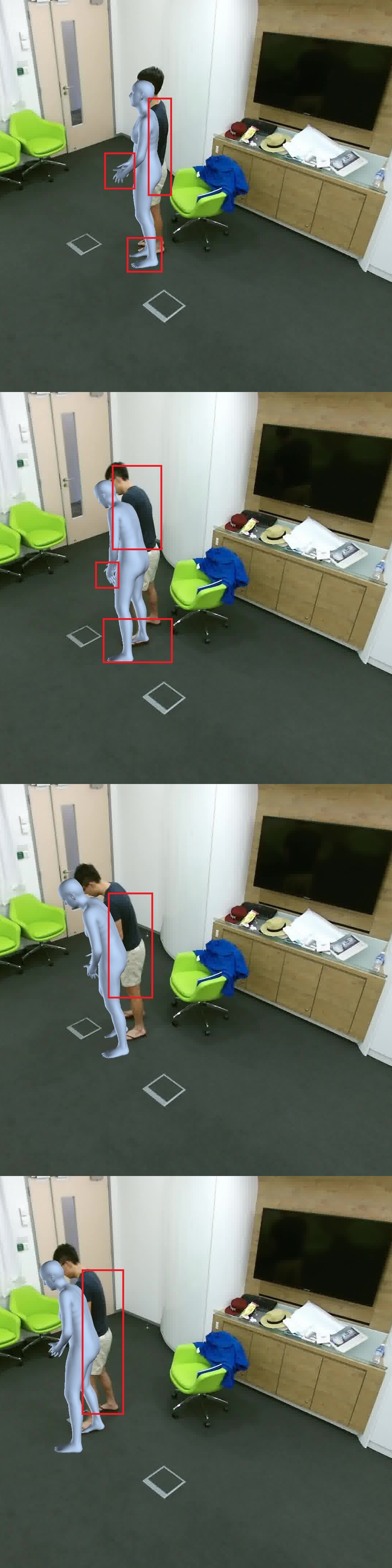}
}
\subfloat[Ours]{
    \label{fig:e}
    \includegraphics[width=0.16\textwidth]{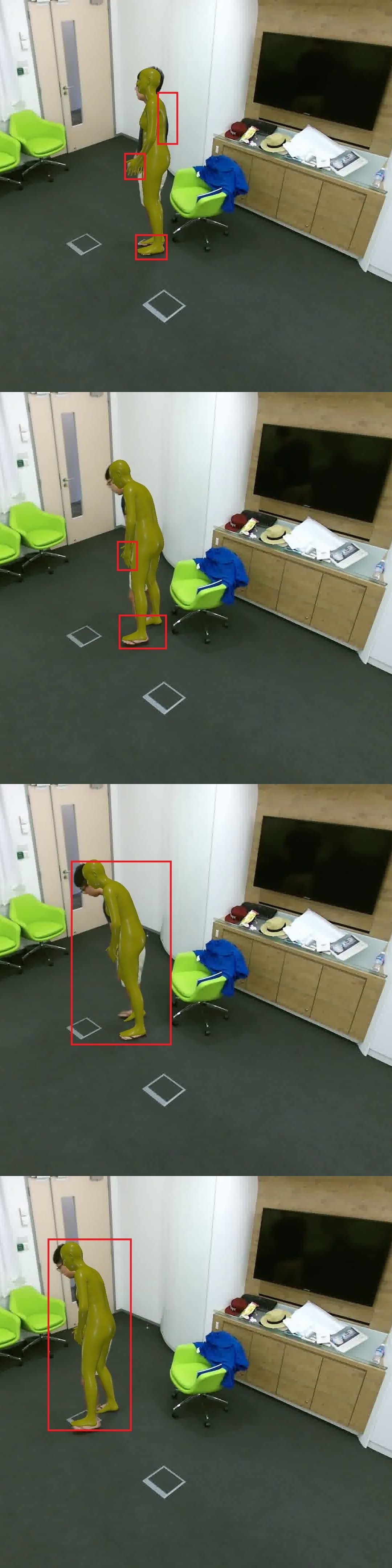}
}
\caption{Qualitative Results on NTU-MC.}
\label{fig:qualitative}
\end{figure*}

\textbf{Qualitative Results. }We have conducted a qualitative evaluation using the Human3.6M, as shown in Fig. \ref{fig:qualitative}. We observe that, in contrast to the HybrIK~\cite{li2021hybrik}, our 3D pose regressor effectively captures information about occluded motions, as highlighted in the red rectangles. Furthermore, we enrich the dataset by incorporating optical flow data, which significantly supplements shape information. Additionally, predicting global trajectories enhances motion information, thereby improving overall robustness. This comparison underlines the strengths of our pose regression method, especially in scenarios involving occlusions.

\subsection{Ablation Studies}
\vspace{-0.2cm}
In order to enhance the performance of the baseline 3D pose regressor under occlusion, we integrate the optical flow and the global trajectory. Therefore, we perform ablation studies to evaluate the contribution of them to the overall pipeline performance. The results of its quantification are shown in Tab. \ref{tab:model_ablation}.

\textbf{Effect on Optical Flow. }Building upon prior research, optical flow data is widely acknowledged as a representation of spatio-temporal feature, demonstrating its utility in 3D pose estimation. Optical flow proves particularly valuable in a multitude of occlusion scenarios—in instances of self-occlusion which can hinder the model's accurate inference of joint positions. Only by utilizing the optical flow information from the initial frames, or from the preceding frame alone, we effectively counteract the disruption caused by self-occlusion, and delivering data augmentation.

In response to this insight, we have designed a two-stream network dedicated to the regression of 3D poses alongside their corresponding 3D mesh parameters, employing no more than the  reverse optical flow from the current frame. 

\textbf{Effect on Global Trajectory. }Recognizing that many approaches to 3D pose estimation overlook the computational significance of the root node, thereby neglecting its critical role in human locomotion analysis, we adopt a global trajectory predictor from GLAMR~\cite{yuan2022glamr}. It facilitates the calculation of the motion direction, velocity, and acceleration pertaining to the joints, enabling us to capture more nuanced and representative motion. 

\begin{table}[t]
\centering
\footnotesize
\begin{tabular}{@{}cccccc@{}}
\toprule
\multirow{2}{*}{Optical flow} & \multirow{2}{*}{Traj pre} & \multicolumn{2}{c}{Human3.6M-TEST} & \multicolumn{2}{c}{3DPW-TEST} \\
\cmidrule(r){3-4} \cmidrule(r){5-6} 
& & PA-MPJPE ($\downarrow$) & {MPJPE} ($\downarrow$) & PA-MPJPE ($\downarrow$) & {MPJPE} ($\downarrow$) \\
\midrule
\checkmark & & 43.9 & 65.2 & 56.5 & 85.5 \\
& \checkmark & 47.6 & 67.7 & 62.3 & 93.0 \\
\checkmark & \checkmark & \bf 37.3 & \bf 58.8 & \bf 49.7 & \bf 80.3 \\
\bottomrule
\end{tabular}
\vspace{0.2cm}
\caption{Quantitative results showing the impact of optical flow and trajectory prediction on pose estimation.}
\label{tab:model_ablation}
\vspace{-1cm}
\end{table}

\section{Conclusion}

 In this paper, we present the OAD2D, an innovative approach for detecting abnormal behaviors (e.g. medical conditions) from 3D motion derived from 2D frames. We reconceptualize abnormal behavior detection by viewing motion as a language, using the Motion-to-Text (M2T) model to transform motion tokens into text tokens for semantic interpretation. A critical step adopted is the use of optical flow to capture 3D motion and pose, which are supplemented by the CVAE and global trajectory optimization to mitigate the occlusion. The integration of a large language model (LLM) enhances our model's generalization in abnormal behavior detection and ability to interpret motion in occluded settings. Our experiments on the NTU-MC and HumanML3D datasets demonstrate promising results. It achieves an accuracy of 85\% and an $F_1$-score of 0.94.

\bibliographystyle{ml_institute}
\bibliography{main}
\end{document}